\documentclass[conference]{IEEEtran}
\IEEEoverridecommandlockouts
\usepackage{cite}
\usepackage{hyperref}
\usepackage{amsmath,amssymb,amsfonts}
\usepackage{algorithm}
\usepackage{algpseudocode}
\usepackage{graphicx}
\usepackage{textcomp}
\usepackage{xcolor}
\usepackage{subcaption}
\def\BibTeX{{\rm B\kern-.05em{\sc i\kern-.025em b}\kern-.08em
    T\kern-.1667em\lower.7ex\hbox{E}\kern-.125emX}}
    
\usepackage{amsmath,amsfonts,bm,amssymb}

\def\eqref#1{equation~\ref{#1}}

\def\1{\bm{1}}

\def\rvh{{\mathbf{h}}}

\def\rvx{{\mathbf{x}}}
\def\rvy{{\mathbf{y}}}

\DeclareMathAlphabet{\mathsfit}{\encodingdefault}{\sfdefault}{m}{sl}
\SetMathAlphabet{\mathsfit}{bold}{\encodingdefault}{\sfdefault}{bx}{n}

\def\sR{{\mathbb{R}}}

\def\sX{{\mathbb{X}}}
\def\sY{{\mathbb{Y}}}

\algrenewcommand\algorithmicrequire{\textbf{Input:}}
\algrenewcommand\algorithmicensure{\textbf{Output:}}

\DeclareMathOperator*{\argmin}{arg\,min}

\newcommand{\ours}{MetaSSD}
\newcommand{\snrdb}{\text{SNR}_{\text{dB}}}

\begin{document}

\title{MetaSSD: Meta-Learned Self-Supervised Detection}

\author{\IEEEauthorblockN{Moon Jeong Park\textsuperscript{1}, Jungseul Ok\textsuperscript{1, 2}, Yo-Seb Jeon\textsuperscript{3}, Dongwoo Kim\textsuperscript{1, 2}}
\IEEEauthorblockA{
\textsuperscript{1}Department of Computer Science, POSTECH \\
\textsuperscript{2}Graduate School of Artificial Intelligence, POSTECH \\
\textsuperscript{3}Department of Electrical Engineering, POSTECH \\
Email: \{mjeongp, jungseul, yoseb.jeon, dongwookim\}@postech.ac.kr
}
}

\maketitle

\begin{abstract}

Deep learning-based symbol detector gains increasing attention due to the simple algorithm design than the traditional model-based algorithms such as Viterbi and BCJR. 
The supervised learning framework is often employed to predict the input symbols, where training symbols are used to train the model.
There are two major limitations in the supervised approaches: a) a model needs to be retrained from scratch when new train symbols come to adapt to a new channel status, and b)
the length of the training symbols needs to be longer than a certain threshold to make the model generalize well on unseen symbols.
To overcome these challenges, we propose a meta-learning-based self-supervised symbol detector named MetaSSD. Our contribution is two-fold: a) meta-learning helps the model adapt to a new channel environment based on experience with various meta-training environments, and b) self-supervised learning helps the model to use relatively less supervision than the previously suggested learning-based detectors.
In experiments, MetaSSD outperforms OFDM-MMSE with noisy channel information and shows comparable results with BCJR. Further ablation studies show the necessity of each component in our framework.


\end{abstract}


\section{Introduction}

Deep learning-based symbol detectors recently gained increasing attention due to relatively simple algorithms than model-based ones~\cite{balatsoukas2019deep, he2019model}. Earlier work in learning-based detectors uses a supervised learning framework to train a model~\cite{shlezinger2020data, shlezinger2020deepsic}. These studies assume that the model parameters are independent of each channel state. Hence, the model parameter needs to be re-estimated whenever a new train signal has arrived.

Online learning-based neural symbol detectors have been proposed to overcome the limitations~\cite{jiang2018artificial, shlezinger2019viterbinet, khani2020adaptive}. These models can adapt to a new environment through an incremental model update via an online learning strategy. Online learning, however, implicitly assumes that the channel state is not changing dramatically, making the models work hard in rapidly changing environments. One can retrain the model parameters from scratch to overcome the limitation, but it is unclear when to retrain the model in such a case.

On the other hand, the learning-based approach often requires more supervision than the model-based approach. Deep neural networks are data-hungry. To make supervised learning stable, one needs to send relatively long train signals, which reduces the efficiency of the communication systems. Moreover, the training time increases as one increases the level of supervision, making the learning-based model unemployable~\cite{khani2020adaptive, he2020model}.

In this work, we propose a new learning framework to address the limitations of the previous work. Our contribution is two-fold: 1) we adopt 
a model-agnostic meta-learning (MAML)~\cite{finn2017model} framework
to find a good initialization parameter, from which the model can adapt to a new environment within a small 
number of update steps, and 2) we propose self-supervised learning that aims to 
predict symbols by minimizing reconstruction error of channel outputs.

The meta-learning helps the model be exposed to various channel environments during meta-training while adapting to a new environment within a few update steps. The self-supervised learning helps the model use relatively less supervision than the previous learning-based detectors.

We compare the performance of our model with BCJR~\cite{bahl1974optimal, li1995optimum} and OFDM-MMSE in a simulated environment. Our approach consistently outperforms OFDM-MMSE and shows comparable or even better results than BCJR when channel estimation is unstable. Further ablation studies on each component of our framework validate the necessity of the proposed framework. 
The code is available at \url{https://github.com/ml-postech/MetaSSD}.



\section{Related Work}

Several deep learning-based symbol detection algorithms are proposed. Among them, a supervised learning algorithm is the most common choice. For example, TISTA~\cite{ito2019trainable} improves stability and performance on sparse signal recovery by adding learnable parameters to the iterative shrinkage thresholding algorithm.
Similarly, \cite{csahin2019doubly}, and \cite{he2020model} improve detection by adding adjustable parameters to the unfolded version of the orthogonal approximate message passing algorithm and expectation propagation algorithm, respectively.
BCJRNet~\cite{shlezinger2020data}, DeepSIC~\cite{shlezinger2020deepsic}, and DLR~\cite{sharma2020deep} perform robust detection when channel estimation is difficult by using deep learning model to learn channel model implicitly.
These models need to be trained from scratch when channel state or SNR changes, making the methods impractical.
VCDN~\cite{samuel2017deep}, TPG-detector~\cite{takabe2019deep}, DetNet~\cite{samuel2019learning}, and CG detector~\cite{wei2020learned} trains the model by the data with various channel state and utilize the channel state as input of the model. They perform detection on changed channels without adaptation at test-time. 
These methods assume that the correct channel state is known, which is not guaranteed in general.

SwitchNet~\cite{jiang2018artificial},  ViterbiNet~\cite{shlezinger2019viterbinet}, and MMNet~\cite{khani2020adaptive} tackle the variability of channel states with an online-learning framework.
SwitchNet trains several models for different channel states offline and learns the parameters online to switch to a suitable model for the current channel state.
ViterbiNet tracks the error rate of the FEC decoder at test time. When the error rate exceeds a predefined threshold, the restored symbols are again used for fine-tuning.
MMNet uses the temporal and spectral locality of the channel model to reduce training time during online learning.
After the channel state of the first subcarrier is learned, it operates as a good initialization point and accelerates training for the rest of the channel information.
Online learning, however, implicitly assumes that the channel state is not changing dramatically, limiting adjustment in a rapidly changing environment.  

EPNet~\cite{zhang2020meta} and Meta-ViterbiNet~\cite{raviv2021meta} adopt a meta-learning framework.
EPNet uses the meta-learned LSTM~\cite{hochreiter1997long} optimizer introduced in \cite{andrychowicz2016learning}.
The meta-learned optimizer is used to learn damping factors sensitive to channel state changes with a small number of epochs. With the proposed damping factor, the optimizer can adapt to the new channel state within a few epochs.
EPNet assumes that the receiver knows the exact channel state. 
Meta-ViterbiNet extends ViterbiNet~\cite{shlezinger2019viterbinet} with MAML.
This work is the most similar to our proposed work. However, Meta-ViterbiNet requires updating the meta-initialization point periodically, increasing the computation cost. 

Note that the number of training symbols for adaptation of the previous methods often exceeds thousands to make the supervised learning stable.
For example, ViterbiNet~\cite{raviv2021meta} and BCJRNet~\cite{shlezinger2020data} use 5,000 and 10,000 symbols for adaptation, respectively, whereas, in our work, we use 100 training symbols, which is significantly less than the previous methods.
\section{Self-Supervised Detector with Meta-Learning}\label{sec:method}

In this section, we propose a Meta-learned Self-Supervised Detector (\ours) to estimate the correct symbols in a finite-memory channel. The proposed model is based on two machine learning frameworks: self-supervised learning and meta-learning.

\begin{figure*}[t!]
    \centering
    \includegraphics[width=0.8\textwidth]{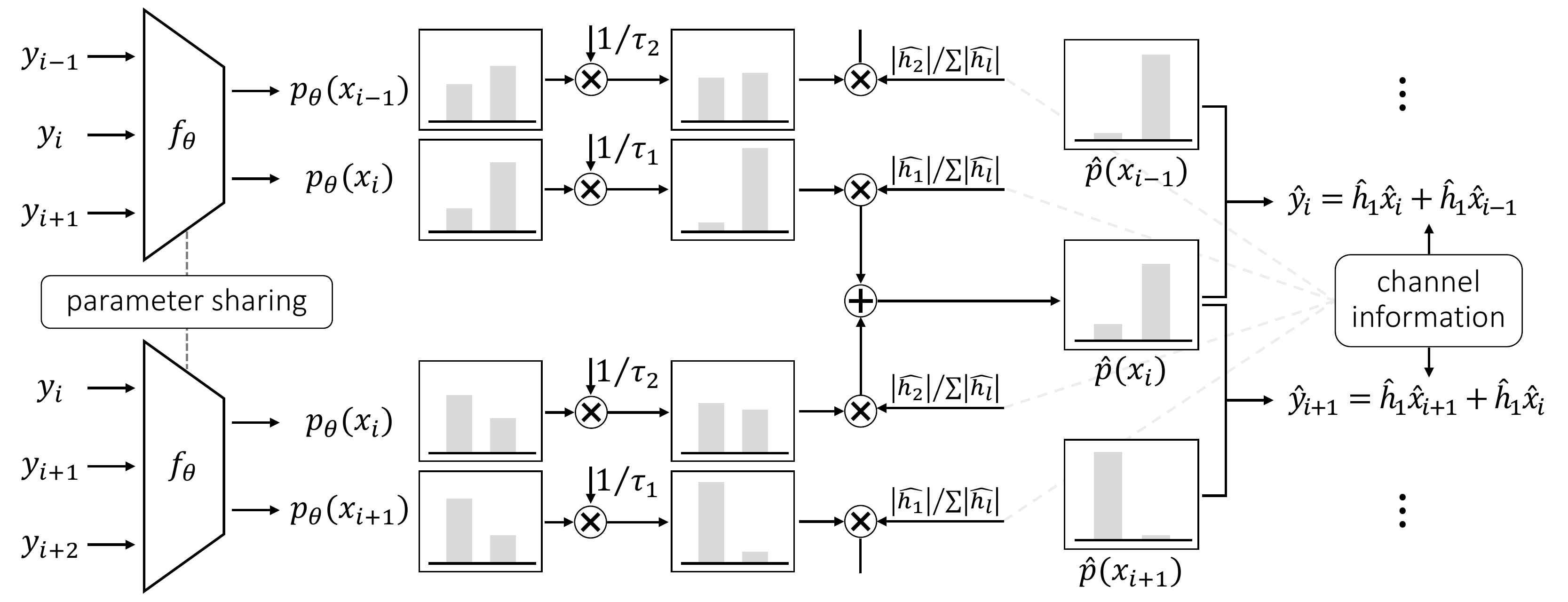}
    \caption{The overall framework of the neural detector with the ISI channel, where we set the memory length to two for the illustrative purpose. The input of the first network is $y_{i-1:i+1}$ and that of second is $y_{i:i+2}$. Both networks share the same parameter $\theta$. As shown in the figure, $x_i$ is predicted twice at a different location of the network. The temperature parameter $\tau_l$ controls the uncertainty of the corresponding location. The weighted sum of two outputs is used as an ensemble prediction, where the weight is proportional to the magnitude $|\hat{h}_l|$ of the corresponding tap.}
    \label{fig:framework}
\end{figure*}

\subsection{Self-supervised neural detector}
We consider a problem of transmitting a block of $N$ symbols over the finite-memory channel. Let $x_i \in \sX$ be a transmitted symbol at time index $i \in [1, 2, ... , N ]$, and $y_i \in \sY$ be a output of the channel at time index $i \in [1, 2, ... , N] $. Given a finite memory length $L$, the conditional distribution of $y_i$ depends on symbols from $i-L+1$ to $i$, i.e., $\rvx_{i-L+1:i}$. 
Therefore, the conditional distribution of outputs $\rvy$ given inputs $\rvx$ can be factorized as
\begin{align}
    \label{eqn:finite-channel}
    p(\rvy | \rvx) = \prod_{i=1}^{N} p(y_i | \rvx_{i-L+1:i}).
\end{align}
If the channel changes over time, the distribution of the outputs also varies over time. 

We aim to design a neural network architecture to recover the original symbols $\rvx$ with a given dataset $D = \{\rvx_{1:P}, \rvy_{1:N} \}$, where $\rvx_{1:P}$ are training symbols and $\rvy_{1:N}$ are channel outputs. We propose a neural network model that predicts a block of $L$ consecutive input symbols $\rvx_{i-L+1:i}$ from a sequence of channel outputs $\rvy_{i-L+1:i+L-1}$. 
Under the assumption that the neural network can approximate an arbitrary relation between the inputs and outputs, we use the channel outputs $\rvy_{i-L+1:i+L-1}$ as an input of the neural network to predict $\rvx_{i-L+1:i}$. 
Specifically, we employ a multi-layered perceptron $f_\theta: \sY^{2L-1} \rightarrow \sR^{|\sX| \times L}$ parameterized by $\theta$ to estimate the distribution over $x_{i-L+1}, \cdots, x_{i}$. Each block of $|\sX|$ outputs are then fed into the softmax layer to estimate the distribution of individual symbol.
In the case of binary symbols, we use $L$ output units with the logistic function to model the distribution over the binary value.
Note that instead of estimating a joint distribution over $\rvx_{i-L+1:i}$, we choose to estimate the probability density of each $x_i$ independently since the output space of the former approach is increasing exponentially with respect to the memory length $L$. 


The proposed neural network predicts $i$th symbol $x_i$ $L$ times at a different position of the network output as the input of the neural network $\rvy_{i-L+1:i+L-1}$ is shifted by one for each prediction.
We predict the same symbol multiple times since a tap with high intensity might be easier to infer than a tap with low intensity.
We then ensemble the estimated distributions to predict $x_i$ with different weights, where the weight is proportional to the estimated intensity of the corresponding tap.
Let $\hat{p}(x_i)$ be the ensembled predicted distribution of $x_i$. We use a cross entropy between the ground truth $\rvx$ and $\hat{p}(x_i)$ to train the model with the following objective function:
\begin{align}
    \label{eqn:supervised_loss}
    \mathcal{L}_{\rvx}(\theta) = \sum_{i=1}^{P}\operatorname{CE}(x_i, \hat{p}(x_i)),
\end{align}
where $P$ is the length of the training symbols, and CE is the cross entropy loss.

When the training length is short, the supervision from the cross entropy would be insufficient to train the model parameter $\theta$. To overcome the limitation, we additionally introduce a self-supervised loss to train the model. Let $\hat{x}_i$ is the ensemble-predicted symbol at time step $i$, i.e., $\hat{x}_i = \sum_{x \in \sX}\hat{p}(x_i = x)x$. We can reconstruct the channel output $y_i$ with a noisy channel information $\hat{\rvh}$. The self-supervised objective can be formalized as
\begin{align}
    \label{eqn:self-sup_loss}
    \mathcal{L}_{\rvy}(\theta) = \sum_{i=1}^{N} \ell(y_i, g_{\hat{\rvh}}(\hat{\rvx}_{i-L+1:i})),
\end{align}
where $\ell$ is a loss function, and $g_{\hat{\rvh}}$ is a channel model-based estimation function of $y_i$ with the channel information $\hat{\rvh}$. For example, with the inter-symbol interference (ISI) channel, the reconstruction can be done via
\begin{align}
    \hat{y_i} = g_{\hat{\rvh}}(\hat{\rvx}_{i-L+1:i}) = \sum_{l=1}^{L} \hat{h}_{l} \hat{x}_{i-l+1},
\end{align} 
where $\hat{y_i}$ is the reconstructed output.
It is worth emphasizing that the self-supervised loss can be applied to each output $y_i$ even without true $x_i$, unlike the supervised loss in (\ref{eqn:supervised_loss}). 

Minimizing the loss in (\ref{eqn:self-sup_loss}) would lead to an incorrect estimation of symbol $\hat{\rvx}$, if the channel information $\hat{\rvh}$ is incorrect. To mitigate the uncertainty of channel information, we additionally introduce a temperature parameter for each tap of the estimated channel state. Let $\tau_l$ is the temperature corresponding to $h_l$. The logits of the output unit corresponding to $h_l$ are then divided by the temperature parameter before the softmax.
Note that the model outputs $|\sX| \times L$ logits before the softmax layer, where each $|\sX|$ logits is used to model the estimated distribution of $l$th location.
Let $[f_\theta(y_{i-L+1:i+L-1})]_{lk}$ be the $k$th output logit of $f_\theta$ at location $l$ given input $y_{i-L+1:i+L-1}$. With temperature, the estimated probability of symbol $p_\theta(x_{i-L+l})$ given $y_{i-L+1:i+L-1}$ can be formalized as
\begin{align}
    p_\theta(x_{i-L+l} = x^{(k)}) = \frac{\exp\left([f_\theta(y_{i-L+1:i+L-1})]_{lk} / \tau_l\right)}{\sum_{k'=1}^{|\sX|} \exp\left([f_\theta(y_{i-L+1:i+L-1})]_{lk'}/ \tau_l\right) },
\end{align}
where $l \in \{1,2,..., L\}$ and $x^{(k)}$ is the $k$th element of $\sX$.
If the temperature $\tau_l$ is close to zero, then the output distribution becomes sparse. If the temperature is high, the output distribution becomes smooth. We fit the temperature parameters with the other parameters while training. Note that the temperature influences the ensembled distribution. 

Finally, we combine the two objectives (\ref{eqn:supervised_loss}) and (\ref{eqn:self-sup_loss}) to train the model:
\begin{align}
    \label{eqn:objective}
    \mathcal{L}_D(\theta) = \mathcal{L}_{\rvx}(\theta) + \alpha \mathcal{L}_{\rvy}(\theta),
\end{align}
where $\alpha$ controls the importance of the self-supervised loss. By minimizing the objective above, we can directly estimate the predicted symbols $\hat{\rvx}$ for the reconstruction of $\rvy$ with $g_{\hat{h}}$. \autoref{fig:framework} illustrates the overall framework of our detector with the ISI channel model.

\subsection{Meta-learning algorithm for neural detector}

In a real-world scenario, the channel state keeps changing over time. So, the detector trained with (\ref{eqn:objective}) needs to be retrained as the new training symbols have been received since the model trained on the previous channel state cannot be generalized to an unseen state. Retraining often requires multiple numbers of iterations over training sets to train a model. Therefore, the real-time responsiveness of the detector cannot be guaranteed with the retraining approach. 

We employ a meta-learning strategy to overcome the problem of model retraining. Recently, meta-learning has emerged as a new framework to learn `how to learn a model'. Specifically, we adopt a MAML~\cite{finn2017model} framework to learn how to make a detector adapt rapidly to a new channel state. The MAML aims to find meta-initialization parameters from which a model can adapt to a new task environment with only a few parameter update steps. During the meta-training step, the model is exposed to multiple tasks, which correspond to different channel realizations in our case. 

Let $\{D^{(t)}\}_{t=1}^T$ be a collection of training symbols and channel outputs, where each dataset consists of training symbols and channel outputs, i.e., $D^{(t)} = \{\rvx_{1:P}^{(t)}, \rvy_{1:N}^{(t)} \}$. We assume that each dataset $D^{(t)}$ is randomly generated from unknown channel distribution $p(\rvh)$ with an independent environmental noise. 
Given meta-training set, MAML aims to find a meta-initialization parameter $\theta^*$ that can minimize the meta-training loss with a single gradient descent step:
\begin{align}
    \theta^* = \argmin_\theta \sum_{t=1}^T \mathcal{L}_{D^{(t)}}(\theta - \lambda \nabla_\theta\mathcal{L}_{D^{(t)}}(\theta)),
\end{align}
where $\lambda$ is a learning rate. Note that the meta-optimization is performed over parameter $\theta$ whereas the objective is computed using the result of the local-optimization with the gradient descent on task $t$.

\begin{algorithm}[t!]
    \caption{Training algorithm for \ours{} \label{alg:maml}}
    \begin{algorithmic}[1]
        \Require task $\{D^{(t)}\}_{t=1}^T$, batch size $B$, adaptation step $K$, learning rate $\lambda, \eta$
        \Ensure meta-initialized $\theta$
        \State Randomly initialize $\theta$
        \While{not done}
            \State Sample batch of tasks $\{D^{(b)}\}_{b=1}^B \subset \{D^{(t)}\}_{t=1}^T$
            \For{$b = 1, 2, \cdots, B$}
                \For{$k = 1, 2, \cdots, K$}
                \State Compute $\theta_k^{(b)}=\theta_{k-1}^{(b)}-\lambda\nabla_{\theta_{k-1}^{(b)}}\mathcal{L}_{D^{(b)}}(\theta_{k-1}^{(b)})$
                \State \Comment ($\theta_0^{(b)} = \theta$)
                \EndFor
            \EndFor
            \State Update $\theta \gets \theta-\eta\nabla_{\theta_{K}^{(b)}}\sum_b\mathcal{L}_{D^{(b)}}(\theta_{K}^{(b)})$
        \EndWhile
    \end{algorithmic}
\end{algorithm}

The meta-optimization can be performed via a stochastic gradient descent as follows:
\begin{align}
    \theta = \theta - \eta \nabla_\theta \sum_{t \in B} \mathcal{L}_{D^{(t)}}(\theta - \lambda \nabla_\theta\mathcal{L}_{D^{(t)}}(\theta)),
\end{align}
where $B$ is a set of randomly sampled mini-batch indices, and $\eta$ is a meta-learning rate. Note that the meta-optimization requires to backpropagate through the Hessian matrix. In practice, the gradient step in local optimization, also called an adaptation step, can be extended to multiple updates, which requires higher-order derivations. In this work, we use the first-order approximation omitting the higher-order derivations~\cite{nichol2018reptile}. The entire algorithm with multiple adaptation steps is presented in \autoref{alg:maml}.

\section{Numerical Experiments}

In this section, we evaluate the performance of the \ours{} proposed in \autoref{sec:method} and compare it with the BCJR and OFDM-MMSE.

\subsection{Experimental Setting}
We conduct experiments with synthetic datasets to evaluate the symbol error rates (SERs) of various detection methods.
We assume the channel state is maintained for $N$ time slots.
We consider an ISI channel where the channel output $y_i$ at time slot $i$ is formalized as:
    \begin{equation}\label{eq:finite_memory_channel}
        y_{i}=\sum_{l=1}^{L} h_{l} x_{i-l+1}+z_{i},
    \end{equation}
where $z_i$ is a noise signal at time slot $i$ distributed as $\mathcal{CN}(0,1/\rho^2)$, and $\rho^2$ represents a signal-to-noise ratio (SNR). 
A BPSK modulator is used, i.e., $\sX=\{-1,1\}$.
The ISI channel is modeled as frequency-selective Rayleigh fading with exponential power delay profile (Exp-PDP), in which the $l$th entry of the ISI channel $\rvh$ is sampled from
\begin{equation}\label{eq:exp-pdp}
    h_{l}\sim\mathcal{CN}(0,\sigma_l^2), 
\end{equation}
with
$$\sigma_l^2=\frac{\exp(-\gamma(l-1))}{\sum_{l=1}^{L}\exp(-\gamma(l-1))},$$ 
where $\gamma$ depends on wireless channel environment. 

\begin{figure}[t!]
    \centering
    \includegraphics[width=0.5\textwidth]{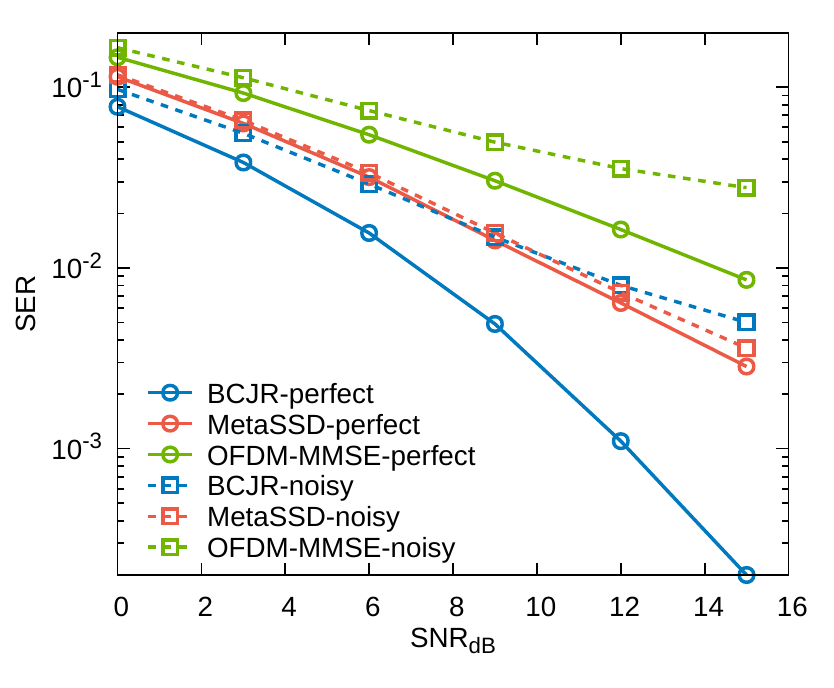}
    \caption{Symbol error rates (SER) on various levels of $\snrdb$. Our model (\ours{}) performs consistently better than OFDM-MMSE. The proposed model shows the least performance changes from the perfect to the noisy environments.}
    \label{fig:main result}
\end{figure}

\begin{figure*}[t!]
    \centering
    \begin{subfigure}[b]{0.32\textwidth}
    \includegraphics[width=\textwidth]{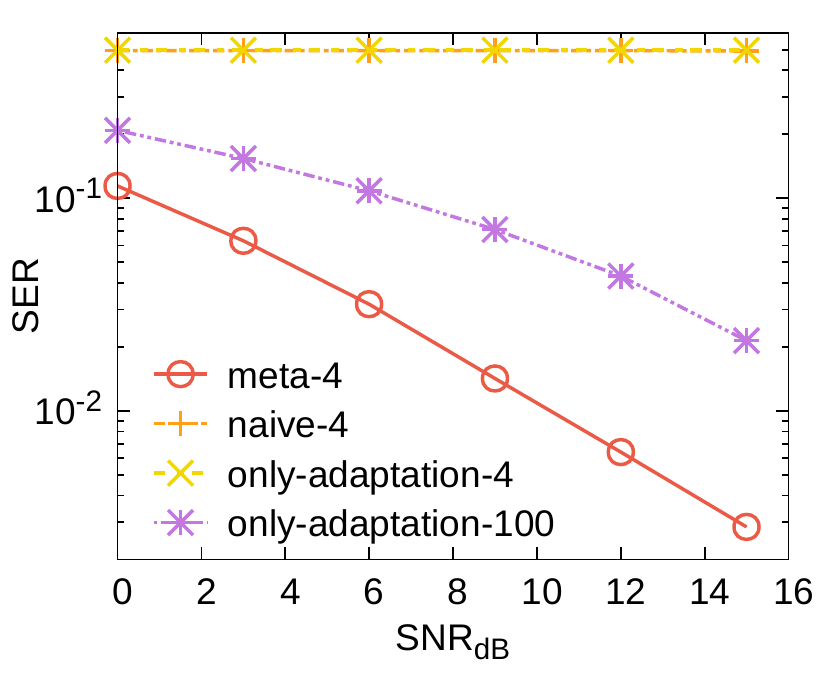}
    \caption{Meta-learning}
    \label{fig:abl-meta}
    \end{subfigure}    
    \begin{subfigure}[b]{0.32\textwidth}
    \includegraphics[width=\textwidth]{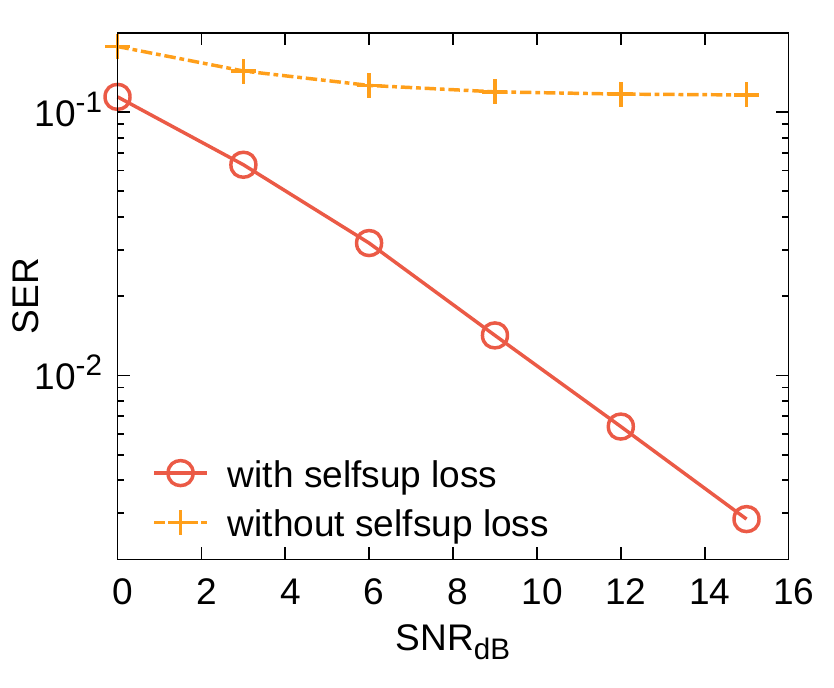}
    \caption{Self-supervised learning}
    \label{fig:abl-ss}
    \end{subfigure}
    \begin{subfigure}[b]{0.32\textwidth}
    \includegraphics[width=\textwidth]{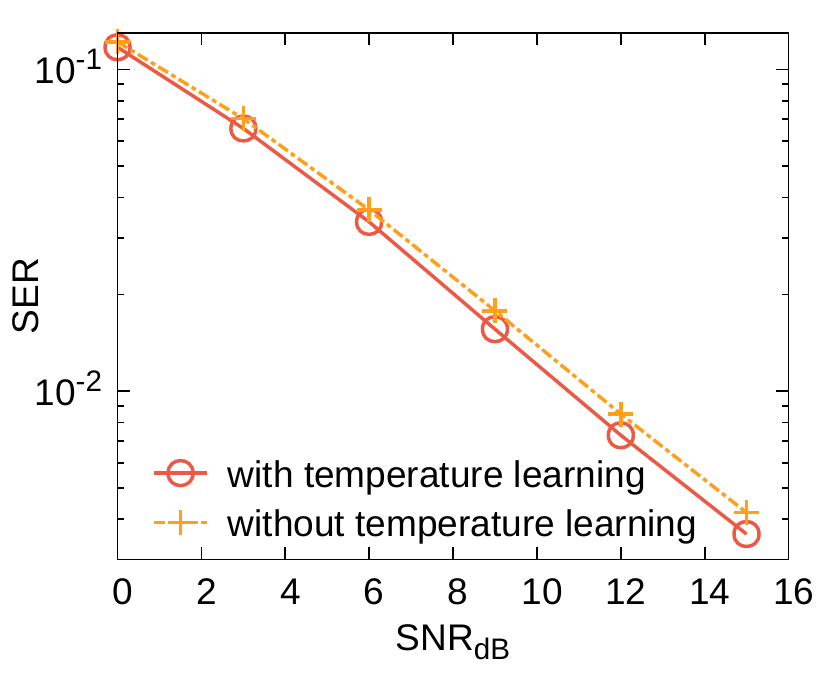}
    \caption{Temperature learning}
    \label{fig:abl-temp}
    \end{subfigure}
    \caption{Ablation study on the three different parts of our learning framework.}
\end{figure*}

A meta-training set $\{D^{(t)}_{\text{train}}\}^T_{t=1}$ is required to train \ours{}. We want to make the model exposed to various environment during training. To do so, we randomly sample $\{h_l\}_{l=1}^L$ from Exp-PDP channel model with $\gamma=2$ and $\snrdb$ from $\text{unif}\{0,15\}$ for each training set $D^{(t)}_{\text{train}}$. 
The size of meta-training set is 10,500, i.e., $T=10,500$, of which $500$ randomly selected sets are used for validation.
Note that the meta-training set consists of various $\snrdb$. Therefore, the meta-initialization parameter obtained from training ensures that the model adapts to the unseen channel regardless of its $\snrdb$.
For the test set, we sampled $500$ datasets for each $\snrdb=\{0,\cdots,15\}$ from the same configuration used in the meta-training. 
In all experiments, we fix the number of training symbols to 100, i.e., $P=100$, memory length to 4, i.e., $L=4$, and the message length to 10,000, i.e., $N=10,000$ for both training and test sets.

A multi-layer perceptron with five-hidden layers is used as $f_\theta$, where 100, 300, 300, 100, and 50 hidden units are used for each layer, respectively. 
ReLu function is used for activation, and the mini-batch size is set to $50$, i.e., $B=50$.
For the local update in the meta-learning, we set the number of adaptation steps to four, i.e., $K=4$.
All hyperparameters, including the learning rate and regularization coefficients, are tuned by the Bayesian optimization~\cite{frazier2018tutorial} method on the validation set.


To compare the robustness of various detection methods against channel information error, we consider the following two scenarios: 
(i) in a \textit{perfect} scenario, we assume that perfect channel information is available at the receiver (i.e., $\hat{\bf h} ={\bf h}$), and (ii) in a \textit{noisy} scenario, we assume that the channel information at the receiver is noisy and modeled as $\hat{\bf h} = {\bf h} + {\bf n}$, where ${\bf n}$ is an additive Gaussian noise whose entry is distributed as $\mathcal{CN}(0,\sigma_n^2)$, as in \cite{shlezinger2020data}.
We assume that $\sigma_n^2 = 0.4$ in our experiments. 

At the test time, we evaluate the SER of \ours{} after performing  $K$ adaptation steps from the meta-initialized parameters for each task $D^{(t)}_{\text{test}}$.


\subsection{Results}

\autoref{fig:main result} shows SER performances of \ours{}, BCJR and OFDM-MMSE.
We average the results of 500 test sets for each $\snrdb$.
The proposed model outperforms OFDM-MMSE consistently across all $\snrdb$ levels for both scenarios.
When the channel information is noisy, \ours{} achieves a lower SER than BCJR for the case of $\snrdb \geq 8$ dB while providing comparable performance for all $\snrdb$ levels.
It is also shown that \ours{} provides the smallest performance gap between \textit{perfect} and \textit{noisy} cases.
This result demonstrates the robustness of \ours{} when the channel information at the receiver is noisy.


\subsection{Ablation Study}

\paragraph{Meta-learning}
To show the effectiveness of a meta-learning framework, we compare two variants of non-meta detection process: \textit{only-adaptation}-$K$ and \textit{naive}-$K$.
\textit{Only-adaptation}-$K$ starts with randomly initialized parameters and updates parameters for $K$ times with the gradient descent algorithm for each test set. 
We note that \textit{only-adaptation}-$K$ is similar to the approach adopted by the previous deep learning-based detector without a meta-learning framework~\cite{ito2019trainable, csahin2019doubly, he2020model, shlezinger2020data, shlezinger2020deepsic, sharma2020deep}.
To compare with the previous work, \textit{only-adaptation}-$K$ uses a relatively less number of training symbols and update steps.
We report the results of two different $K$: 4 and 100.
\textit{Naive}-$K$ predicts symbols with the initialization parameter obtained by \autoref{alg:maml} without any local update step.
\textit{Naive}-$K$ assumes all meta-training data are drawn from the same distribution corresponding to the standard supervised learning. 
At test time, \textit{naive}-$K$ performs $K$ number of local update steps for each test set.

The comparison between meta and non-meta strategies in \textit{perfect} scenario is shown in \autoref{fig:abl-meta}. \textit{meta}-$K$ denotes our model.
Relatively low performance of \textit{naive}-$4$ highlights the importance of the local adaptation step in training. 
We observe that accuracy of \textit{meta}-$4$ is even higher than \textit{only-adaptation}-$100$. The \textit{only-adaptation}-$100$ fails to achieve comparable performance.
The result 
demonstrates the importance of the meta-initialization parameter trained by the meta dataset.

\paragraph{Self-supervised learning}
To verify the effectiveness of our self-supervised learning framework, we 
train the model without self-supervised loss. 
\autoref{fig:abl-ss} compares the difference between the presence and absence of self-supervised loss. All models are trained with the meta-learning algorithm.

We observe that the model trained with the self-supervised loss performs significantly better in all $\snrdb$ levels. The results show that the 100 training symbols are insufficient to train a model even with the meta-initialization and highlight the importance of the self-supervised loss.

\paragraph{Temperature learning}
We study the influence of the learnable temperature parameter $\tau$ in the softmax function.
We compare the performance of the softmax with and without temperature.
Since the learnable temperature is designed to reduce the negative effect of the channel estimation noise, we conduct a comparison in \textit{noisy} environment.
\autoref{fig:abl-temp} shows the result.
Although the improvement seems marginal, the model with the learnable temperature consistently outperforms the ones without the temperature across all $\snrdb$ levels.

\section{Conclusion}

In this work, we propose a meta-learning and self-supervised learning-based symbol detection algorithm. Unlike the existing work on deep learning detectors, the proposed model can be adapted to a new channel environment with a fixed number of adaptation steps while enjoying relatively less supervision on training signals. The experimental results show that the proposed model outperforms OFDM-MMSE and shows comparable performance with the BCJR algorithm when the channel information at a receiver is noisy.


\newpage

\bibliographystyle{IEEEtran}
\bibliography{references}

\end{document}